\documentclass[10pt, a4paper]{article}

\usepackage{placeins}

\usepackage{listings}
\usepackage[final]{lrec2026}

\title{A Human-in/on-the-Loop Framework for Accessible Text Generation}

\name{Lourdes Moreno, Paloma Martínez}

\address{Universidad Carlos III de Madrid\\
Av. Universidad, 30, Leganés, 28911, Spain\\
lourdes.moreno@uc3m.es, paloma.martinez@uc3m.es
}

\abstract{
Plain Language and Easy-to-Read formats in text simplification are essential for cognitive accessibility. Yet current automatic simplification and evaluation pipelines remain largely automated, metric-driven, and fail to reflect user comprehension or normative standards. This paper introduces a hybrid framework that explicitly integrates human participation into LLM-based accessible text generation. Human-in-the-Loop (HiTL) contributions guide adjustments during generation, while Human-on-the-Loop (HoTL) supervision ensures systematic post-generation review. Empirical evidence from user studies and annotated resources is operationalized into (i) checklists aligned with standards, (ii) Event-Condition-Action trigger rules for activating expert oversight, and (iii) accessibility Key Performance Indicators (KPIs). The framework shows how human-centered mechanisms can be encoded for evaluation and reused to provide structured feedback that improves model adaptation. By embedding the human role in both generation and supervision, it establishes a traceable, reproducible, and auditable process for creating and evaluating accessible texts. In doing so, it integrates explainability and ethical accountability as core design principles, contributing to more transparent and inclusive NLP systems.
 \\ \newline }

\begin{document}

\maketitleabstract

\section{Introduction}

Access to understandable information is a fundamental right, particularly in critical domains such as health, education, legal, and public administration. Many people face barriers when interacting with complex or technical texts, including people with low literacy, reading comprehension difficulties, cognitive impairments, older adults, and non-native speakers \citep{OECD2024}. To address these challenges, two accessibility-oriented content formats have gained prominence: Plain Language (PL), codified in ISO 24495-1:2023 \citep{ISO24495_1_2023}, and Easy-to-Read (ER), defined by standards such as the Spanish standard UNE 153101:2018 \citep{UNE153101_2018} and international guidelines \citep{IFLA2010,InclusionEurope2025}.

Recent advances in Natural Language Processing (NLP), particularly with Large Language Models (LLMs), have enabled automatic text simplification at lexical, syntactic, and discourse levels \citep{Vaswani2017,Nisioi2017,Shardlow2024}. Nevertheless, evaluating and ensuring the accessibility of such outputs remains a major challenge. Widely used metrics such as BLEU \citep{Papineni2002}, SARI \citep{Xu2016}, SAMSA \citep{Sulem2018}, or BERTScore \citep{Zhang2019} capture partial aspects of quality-lexical simplicity, structural fidelity, and semantic similarity-but fail to reflect user comprehension \citep{AlvaManchego2021b}.

Parallel to advances in NLP, the Human-Computer Interaction (HCI) community has emphasized participatory and empirical validation methods for accessibility \citep{Lazar2017}. Research described in \citep{Moreno2024UAIS,Alarcon2024EASIERSystemIJHCI,Alarcon2023EasierCorpusPLOS} combines NLP resources with user-centered evaluations providing both methodological insights and empirical evidence on how cognitive accessibility can be implemented in practice.

However, most current approaches rely either on automatic metrics or ad hoc human validation, rarely integrating both into a systematic workflow. In the field of NLP and human-AI interaction, this integration has been conceptualized through Human-in-the-Loop (HiTL) and Human-on-the-Loop (HoTL) approaches. HiTL involves human input during the training or generation process, e.g., selecting or editing candidate simplifications, or guiding the model in fine-tuning \citep{Amershi2014,Zanzotto2019}. HoTL, by contrast, emphasizes post-hoc supervision, where humans oversee system outputs through structured review, guidelines, or monitoring mechanisms \citep{Shneiderman2020}. While HiTL has been applied to tasks such as interactive summarization or controllable text generation, and HoTL to safety monitoring in AI systems, these paradigms have not yet been fully exploited for cognitive accessibility in text simplification.

To fill the gap between automatic simplification and actual cognitive accessibility, we propose a framework that makes human participation a core design principle. HiTL enables interactive guidance during generation, such as accepting or rejecting lexical substitutions, while HoTL supervision provides structured oversight through checklists, trigger rules, and accessibility key performance indicators (KPIs) after outputs are produced. Together, these mechanisms ensure that accessible text generation is not only driven by metrics but also grounded in human judgments, standards, and empirical validation.

At a glance, the framework combines: (i) HiTL guidance during generation, (ii) automatic metric-based checks, (iii) Event--Condition--Action (ECA) triggers for escalation, (iv) HoTL expert review, and (v) adaptation through structured feedback.

The remainder of this paper is organized as follows: Section~2 reviews related work on language accessibility standards, Section~3 reviews previous work on text simplification evaluation and HCI methods, and Section~4 presents background resources and empirical evidence. Section~5 introduces the proposed HiTL/HoTL framework (workflow, checklist, triggers, KPIs, integration). Section~6 discusses ethical and governance issues, and Section~7 concludes with a discussion and future work.

\section{Plain Language (PL) and Easy-to-Read (ER) Standards}

For NLP tools to be effective and socially relevant, they must be aligned with PL and ER standards, which promote the adaptation of texts to the real needs of their users, especially those with intellectual disabilities (ID) or reading limitations.

The main international reference is ISO 24495-1:2023, which defines general principles applicable to any language and domain, establishing that texts must be easy to find, understand, use, and evaluate by their target audience \citep{ISO24495_1_2023}.

At the European level, the European Strategy on the Rights of People with Disabilities 2021--2030 highlights the importance of access to understandable information for persons with disabilities, recognizing ER as a key tool for achieving cognitive accessibility. Similarly, Directive (EU) 2016/2102 requires public sector websites and mobile applications to be accessible in accordance with standards such as the European standard EN 301 549 \citep{EN301549_2021} and the WCAG web content standard \citep{WCAG_W3C}, which includes requirements on textual comprehensibility.

In Spain, the UNE 153101 EX:2018 standard on Lectura F{\'a}cil (LF) stands out, establishing guidelines for the drafting, layout, and validation of texts, emphasizing the participation of people with ID in the evaluation process\citep{UNE153101_2018}. There are other national standards and legal frameworks that also shape PL/ER practice. Examples include {\"O}NORM V2100:2017 in Austria (administrative texts in LF), DIN SPEC 33429 in Germany (PL criteria for public documents), and the Equality Act (2010) in the United Kingdom, which requires public and private organizations to ensure accessibility, including clear information for people with cognitive disabilities. In addition, the UK General Data Protection Regulation (UK GDPR) and the Data Protection Act (2018) require privacy-related information to be clear and understandable. In healthcare, the NHS England Accessible Information Standard (AIS) requires accessible formats such as ER for people with cognitive or sensory disabilities. The Plain English Campaign certifies texts with the Crystal Mark, a symbol of linguistic clarity, while the GOV.UK Style Guide mandates the use of clear language on government digital platforms \citep{PLAIN2021}.

Although PL and ER share the overall goal of making text understandable, there are important differences. PL is aimed at a broad audience and seeks to reduce unnecessary technical terms and complex expressions. It typically relies on brevity, inclusive everyday language, and a structure that allows key information to be easily located, using direct style, active voice, and short paragraphs. ER is specifically designed for people with reading comprehension difficulties, such as people with ID, and therefore requires stricter guidelines for simplification and content presentation: bulleted lists, high-frequency words, abundant use of visual examples, and single-idea sentences. In addition, verification with real readers with ID is essential to validate ER texts \citep{IFLA2010,InclusionEurope2025,UNE153101_2018}.

The core principles shared by PL and ER include avoiding information overload, using familiar examples and metaphors, maintaining consistency in terminology, and reinforcing content with visual elements that facilitate comprehension. These guidelines constitute a valuable source of best practices that NLP developers can implement in automatic text simplification systems: controlling sentence length, adapting style to the target audience, and validating outputs with real users to ensure that simplifications effectively improve comprehension, usability, and cognitive accessibility.

\section{Related Work}

Research on text simplification and accessibility has evolved along three main lines: automatic evaluation metrics, HCI methods for cognitive accessibility, and human participation in NLP workflows. These strands converge in the challenge of ensuring that automatically simplified outputs are not only linguistically well-formed, but also cognitively accessible and aligned with normative standards.

Early work on automatic metrics focused on surface-level overlap with reference corpora, such as BLEU \citep{Papineni2002} and ROUGE \citep{Lin2004}. While useful in machine translation, these metrics have drawbacks as they penalize valid reformulations and ignore structural or semantic changes \citep{Xu2016,Sulem2018}. Dedicated metrics such as SARI \citep{Xu2016} and DSARI \citep{Maddela-Controllable-2021} improved lexical coverage by rewarding additions, deletions, and retentions, but still fail to capture global readability and meaning preservation. Structural metrics like SAMSA \citep{Sulem2018} focus on sentence splitting and discourse structure, while semantic similarity methods such as BERTScore \citep{Zhang2019} and MoverScore \citep{Zhao2019} leverage embeddings to better capture meaning preservation. More recently, LENS \citep{Maddela2022} has demonstrated a higher correlation with human judgments by learning evaluation from annotated data. However, systematic reviews conclude that no single metric fully captures the multifaceted requirements of accessibility \citep{AlvaManchego2021b}.

In parallel, the HCI community has developed methods to assess and design for cognitive accessibility. Standards \citep{ISO24495_1_2023,COGA2021_W3C,WCAG22_W3C,EN301549_2021} emphasize clarity, structure, and validation with end users. Empirical studies show that accessible interfaces can enhance comprehension when supported with contextual glossaries, pictograms, and multimodal cues. \citet{Alarcon2024EASIERSystemIJHCI,Moreno2024UAIS} developed an approach that combined corpus-based resources with participatory evaluations, validating glossary-based user interfaces with users with diverse cognitive profiles and developing annotated corpora for lexical simplification, \citep{Alarcon2023EasierCorpusPLOS}. This line of work illustrates how user-centered design and linguistic resources can be integrated into accessibility evaluation. Works including user evaluations of simplification approaches \citep{Ledoyen2025} proposed an evaluation framework in which human assessments supplement automatic metrics. \citet{Goldsack2023} evaluated a system that generates summaries of biomedical articles for lay readers, focusing on readability and factuality; the evaluation was conducted by NLP experts through questionnaires, leaving lay users aside. Other approaches simulate human behavior using multi-agent systems, such as ExpertEase \citep{Mo2024}, where multiple agents cooperate on text simplification tasks using external tools and guidelines, but no human evaluation is performed.

Finally, recent work has explored human participation in NLP architectures. HiTL approaches are widely used in annotation and training pipelines, allowing humans to refine data, correct model outputs, or steer generation in real time \citep{Chiang2023}. \citet{Gao2025} describe an evaluation of an LLM-based simplification system aligned using Direct Preference Optimization (DPO), using human feedback from people with ID (15 participants) and expert evaluators. The final evaluation indicated the effectiveness of DPO for personalizing LLM-based simplification, but also points to challenges such as improved HCI methods for eliciting preferences. More recently, \citet{Ilgen2025} defined a human-centered readability score based on a five-dimensional framework (tone, trust, cultural relevance, feasibility, and clarity), integrating automatic metrics with human experience around readability.

HoTL approaches emphasize structured supervision and post-hoc control, in which humans monitor, validate, or adjust system outputs through review procedures, guidelines, or oversight mechanisms \citep{Li2020HOTL,Shneiderman2020,Carrer2024}. In the context of text simplification, these paradigms remain underexplored, although related work in summarization and factual consistency shows the potential of combining automatic scoring with human oversight \citep{Manakul-2023}. This convergence reinforces the shift toward hybrid workflows that systematically integrate human validation within NLP evaluation pipelines.

Recent work has also addressed the trustworthiness and safety of text simplification. \citet{Hayakawa2025} proposed an efficient framework for lexical simplification with small LLMs, emphasizing privacy, efficiency, and the detection of harmful simplifications through probabilistic safety signals. Their approach highlights the need for traceable and accountable simplification pipelines.

In summary, existing metrics, HCI methods, and human-in/on-the-loop frameworks each addressing parts of the problem but rarely intersecting. What is missing is a unified approach that reuses language resources, incorporates HCI evidence, and defines hybrid workflows where humans complement automation in the generation, evaluation, and adjustment of PL and ER outputs. This gap motivates the approach proposed in this paper.

\section{Background Resources}

Evidence is drawn from user studies and corpora developed in previous research on cognitive accessibility. Data from older adults, people with intellectual disabilities (ID), and expert annotators are used in tasks such as complex word identification (CWI), synonym acceptance, corpus annotation, and User Interface (UI) validation. Only selected findings are summarized here, providing preliminary evidence to motivate a hybrid HiTL-HoTL framework and serving as the basis for accessibility KPIs, review triggers, and human intervention points.

\subsection{Complex Word Identification Task}

\citet{Alarcon2024EASIERSystemIJHCI} did an evaluation for CWI task where 50 participants (25 older adults and 25 with ID) were asked to highlight words considered difficult in domain-specific texts. In this study, participants directly marked the words they perceived as difficult, and system behavior was analyzed through standard CWI metrics as user-grounded evidence of perceived lexical difficulty.

Results indicated systematic differences between groups. Older adults tended to select a larger number of candidates, which led to higher recall (0.73) but lower precision (0.54). By contrast, participants with ID marked fewer items, resulting in higher precision (0.58) and a lower F1 score (0.56).

Within each group, educational background also influenced performance: older adults with higher education showed a better balance between recall and precision, whereas participants with lower literacy showed greater variability in their selections. These complementary profiles suggest that while older adults provide broader coverage, participants with ID contribute more selective judgments. This divergence motivates adaptive HiTL mechanisms that can balance broader lexical coverage against more selective judgments across user profiles.

\subsection{Synonym Acceptance Task}

\citet{Alarcon2024EASIERSystemIJHCI} carried out an experiment with the same 50 participants (25 older adults and 25 with ID), who were asked to judge the adequacy of proposed synonyms to complex words in the context. Results indicate profile-dependent differences across groups (Table~\ref{tab:syn-accept}). Older adults tended to reject all candidates more frequently (16.6\%) than participants with ID (7.6\%). In contrast, participants with ID showed a stronger tendency to accept at least one synonym (67.5\% vs \ 58.9\%),  while full acceptance was very similar across groups (24.5\% vs \ 24.9\%). These patterns suggest that participants with ID may be more flexible in their judgments, whereas older adults are more conservative.

Beyond group-level differences, educational background also influences synonym acceptance. Participants with lower education levels (primary or no studies) achieved higher rates of correct acceptance ($\approx 78$--$80\%$), while those with higher education were more likely to partially accept or reject candidates (e.g., $\approx 65\%$ correct with secondary education). This effect was statistically significant, confirming that literacy levels shape how simplifications are perceived and judged.

Taken together, these results indicate that synonym acceptance is not homogeneous across user profiles, and that both group characteristics (older vs.\ ID) and individual factors (such as education level) must be considered when defining acceptance thresholds. For HiTL workflows, this evidence motivates adaptive mechanisms sensitive to user profile; for HoTL supervision, it suggests the need to monitor systematic rejection patterns as triggers for expert review.
\begin{table}[!t]
\centering
\small
\setlength{\tabcolsep}{5pt}
\renewcommand{\arraystretch}{1.1}
\begin{tabular}{lccc}
\hline
\textbf{Group} & \textbf{None} & \textbf{Some} & \textbf{All} \\
\hline
Older adults & 16.6\% & 58.9\% & 24.5\% \\
Participants with ID & 7.6\% & 67.5\% & 24.9\% \\
\hline
\end{tabular}
\caption{Synonym acceptance rates by participant group ($n=50$). \textit{None/Some/All} = none accepted / some accepted / all accepted.}
\label{tab:syn-accept}
\end{table}

\subsection{Expert Annotation Reliability}

Concerning resources, EASIER corpus, \citep{Alarcon2023EasierCorpusPLOS}, was annotated by three expert linguists and is composed of 260 documents with 8{,}155 words labelled as complex words and 5{,}130 words with at least one proposed context-aware synonym associated. This corpus provides evidence about the reliability of lexical complexity judgments. Agreement was higher for multi-word expressions ($\kappa = 0.64$) than for single words ($\kappa = 0.52$), suggesting that multi-words may be more stable units for supervised training and for HoTL supervision. These results provide complementary evidence to user studies and support the integration of supervised resources into hybrid evaluation workflows.

\subsection{HCI-based Validation Methods}

In addition to corpus annotation, \citep{Moreno2024UAIS} describes a system where glossaries of definitions, synonyms, and pictograms are integrated into cognitively accessible user interfaces, which were validated with end users, including older adults and people with ID. Validation relied on HCI-based methods applied directly with participants \citep{Lazar2017}, such as: (i) comprehension questionnaires measuring retention of key information; (ii) navigation tasks evaluating time and success rates to locate content in documents; (iii) glossary activation logs recording actual use of contextual support; and (iv) usability feedback collected through standardized instruments. These methods provided empirical evidence of how accessibility is perceived and implemented by real users, beyond what automatic metrics can analyze.

\subsection{Resources for Language Model Adaptation}

Beyond user studies and user interface validation, evidence also comes from resources applied in simplification system development in evaluation campaigns. Participation in a shared task on text simplification \citep{Moreno2025CLEARS} demonstrated that widely used automatic simplification metrics (e.g., BLEU, SARI, FKGL) fail to capture comprehension and accessibility principles, requiring complementary human feedback for calibration. In parallel, \citep{Martinez2024FrontiersE2R} emphasized the need to align evaluation with international standards such as ISO 24495-1 and with empirical user evidence.

These works highlight two key points. First, resources such as annotated corpora, synonym judgments, and comprehension tests are not only valuable for benchmarking but can also inform prompt design and fine-tuning of LLMs. For instance, simplification rules derived from corpus annotation can be encoded as constraints in prompting strategies. Second, iterative metrics-based evaluation cycles followed by human review and adjustment constitute an implicit HiTL-HoTL workflow, showing how human participation can guide and supervise LLM adaptation.

These resources also anticipate integration into adaptive training pipelines in the future. Preference signals such as synonym acceptance thresholds, comprehension test scores, and glossary activation logs can be repurposed not only to evaluate outputs but also to supervise iterative fine-tuning cycles. This perspective directly links empirical evidence with emerging preference-based approaches to LLM alignment. In this sense, accessibility-grounded corpora provide domain-specific signals that can complement large-scale, generic datasets in both prompting strategies and preference-based adaptation.
\subsection{Implications for HiTL--HoTL Workflows}

Taken together, these results illustrate how human participation most effectively complements automation. Older adults provide coverage in CWI, people with ID contribute precision, expert annotation stabilizes supervision via multi-word units, and validated interface patterns ensure delivery through accessible modalities. Moreover, resources for model adaptation \citep{Martinez2024FrontiersE2R} and evidence of metric limitations \citep{AlvaManchego2021b,Moreno2025CLEARS} indicate that human input must not only validate but also guide training and prompting strategies.

This convergence of corpus-based evidence, user studies, and model adaptation resources sets the stage for a hybrid framework. Section~5 implements these insights by defining a Human-in/on-the-Loop (HiTL--HoTL) workflow structured around checklists, trigger rules, and KPIs.

\section{Proposed Framework}

\begin{figure*}[t]
  \centering
  \includegraphics[width=\textwidth]{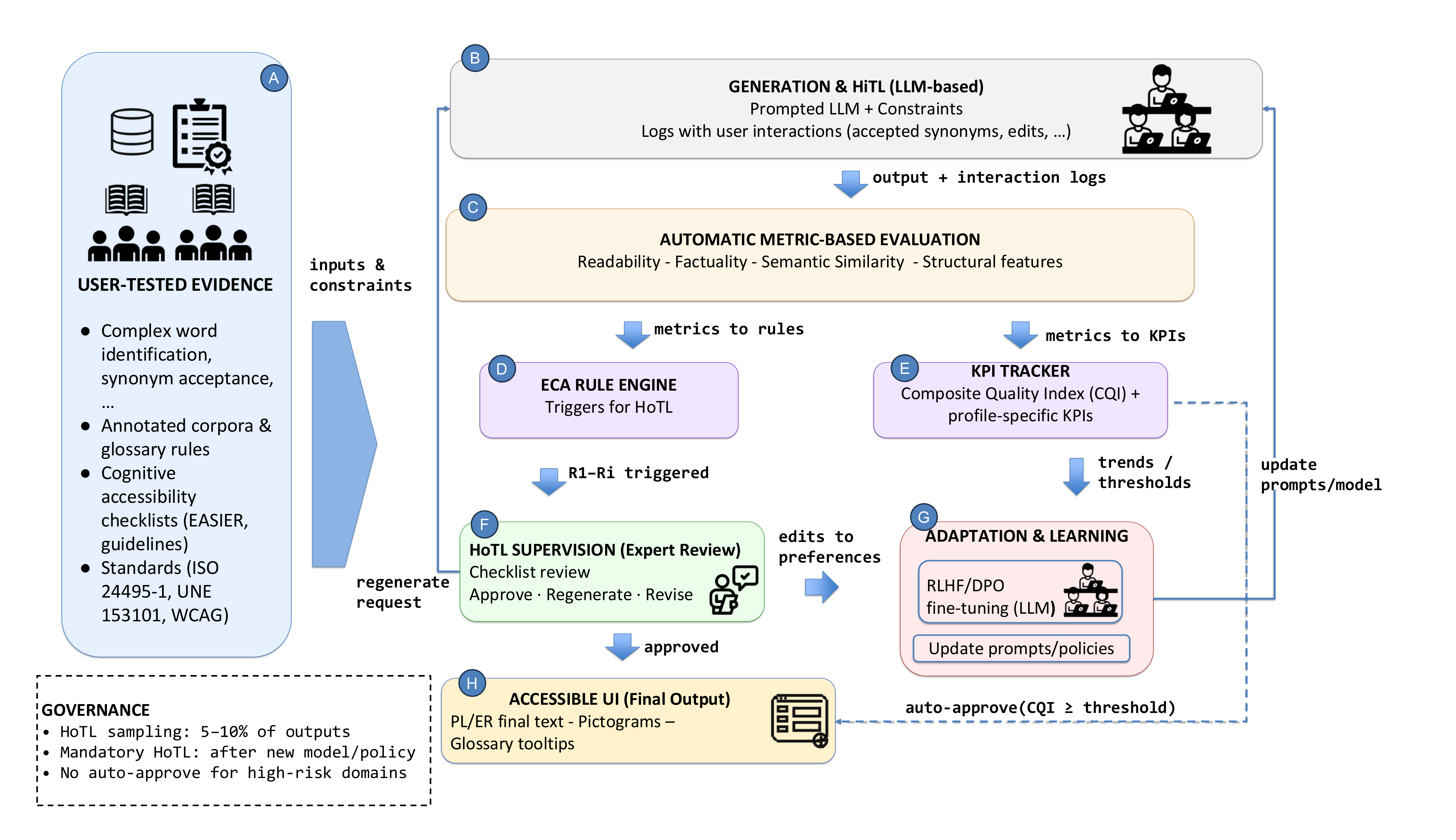}
  \caption{Overall workflow of the proposed HiTL--HoTL framework.}
  \label{fig:workflow}
\end{figure*}

We propose a hybrid framework that integrates Human-in-the-Loop (HiTL) and Human-on-the-Loop (HoTL) mechanisms across accessible text generation, metric-based evaluation, and adaptation of PL/ER outputs. Grounded in the evidence summarized in Section~4 (CWI, synonym acceptance, expert annotation, UI validation, and model adaptation), the framework operationalizes this evidence through checklists, ECA trigger rules, and profile-specific KPIs that govern the workflow depicted in Figure~\ref{fig:workflow}. The contribution is therefore an operational and traceable framework that consolidates prior empirical evidence, annotated resources, and accessibility-oriented supervision into a unified workflow.

\subsection{Overall Workflow}

At a high level (Figure~\ref{fig:workflow}), the workflow is a multi-stage, iterative pipeline. Evidence and resources (A) condition generation with HiTL (B) through inputs and constraints; the output plus interaction logs are then automatically evaluated (C). From (C), metrics feed rule-based checking (D) and quality indicators (E) that determine routing: if rule conditions are met, the  ECA rule engine activates HoTL supervision (F); if quality trends or thresholds suggest systematic issues, signals are sent to Adaptation and Learning (G). Expert editing operations from (F) are consolidated in (G) and returned to (B) as updates to prompts/policies or the model, closing the loop; once an approved decision is issued in (F), the result is delivered via an accessible User Interface (H).

Evidence from Section~4 indicates that differences in CWI between older adults and people with ID inform HiTL checkpoints; profile-specific synonym-acceptance patterns motivate HoTL escalation; higher agreement in multi-word annotations defines stable units for supervision; and user-validated resources can be embedded into prompts and fine-tuned models for adaptation.

The following sections provide the technical details: Section~5.2 on HiTL adjustments, Section~5.3 on the HoTL checklist, Section~5.4 on rule-based routing, Section~5.5 on quality indicators, and Section~5.6 on how these signals are leveraged for adaptation.

\subsection{Human-in-the-Loop (HiTL) Adjustments}

HiTL participation is enacted during generation (Figure~\ref{fig:workflow}, B). Evidence from Section~4 indicates broader coverage (higher recall) for older adults and higher precision for participants with ID; these complementary profiles are implemented through interactive steps that (i) request short, active sentences, (ii) discourage jargon and expand acronyms, (iii) split long sentences, and (iv) fix meaning distortions while preserving facts and numeric integrity. User-validated resources (Section~4), annotation guidelines, synonym judgments, and glossary constraints are encoded directly into prompts or, when needed, into fine-tuning procedures. Interaction traces, produced at generation time, feed automatic evaluation (Figure~\ref{fig:workflow}, B$\rightarrow$C), while stabilized guidance (e.g., glossary rules) is later consolidated in adaptation and learning and pushed back to generation (Figure~\ref{fig:workflow}, G$\rightarrow$B).

\subsection{Human-on-the-Loop (HoTL) Checklist}

While HiTL constrains and guides generation under uncertainty, HoTL provides accountable human adjudication when borderline, risky, or conflicting cases cannot be resolved with sufficient confidence through automatic checks.

HoTL supervision provides a structured review layer when automatic metrics do not guarantee comprehension. The checklist integrates international standards \citep{ISO24495_1_2023,UNE153101_2018} with evidence from synonym acceptance, expert-annotation reliability, interface validation, and model-adaptation studies (Section~4).

Table~\ref{tab:hotl-checklist} summarizes how accessibility dimensions map to concrete checklist items with human-centered validation (e.g., comprehension tests, reading times, usability questionnaires), including a row on prompt/model adaptation where user-validated resources (corpus rules, glossary constraints, synonym judgments) can be encoded into prompt design or fine-tuning.

\begin{table*}[t]
\centering
\small
\setlength{\tabcolsep}{6pt}
\renewcommand{\arraystretch}{1.15}
\begin{tabular}{p{0.18\textwidth} p{0.47\textwidth} p{0.30\textwidth}}
\hline
\textbf{Dimension} & \textbf{Operational check} & \textbf{Human-centered validation} \\
\hline
Lexical clarity & Common words; explain acronyms & Comprehension tests \\
Syntactic simplicity & $\leq$20 words per sentence; one idea per sentence & Eye-tracking; reading times \\
Structural clarity & Logical order; headings; lists & Task success rates \\
Relevance & Preserve essentials; avoid redundancy & User judgments \\
Multimodal support & Glossaries; pictograms & Usability questionnaires; user testing \\
Prompt/model adaptation & Encode rules, glossary constraints, and domain specificity & HiTL prompt design; fine-tuning \\
\hline
\end{tabular}
\caption{HoTL checklist integrating accessibility dimensions with human-centered validation.}
\label{tab:hotl-checklist}
\end{table*}


Outputs are considered checklist-compliant when at least two-thirds of the dimensions are satisfied. In the workflow, expert review is reached when rule checks route an item to HoTL (Figure~\ref{fig:workflow}, D$\rightarrow$F); edits inform adaptation for subsequent cycles (Figure~\ref{fig:workflow}, F$\rightarrow$G), regeneration can be requested where appropriate (Figure~\ref{fig:workflow}, F$\rightarrow$B), and delivery follows an approved decision (Figure~\ref{fig:workflow}, F$\rightarrow$H).

\subsection{Trigger Rules for Review}

Systematic synonym-rejection patterns (Section~4) and divergences between readability and comprehension (Section~4) motivate ECA rules that determine when HoTL supervision is required. Following automatic evaluation (Figure~\ref{fig:workflow}, C), metric outputs feed the rule engine; when trigger conditions are met, cases are escalated to expert review (Figure~\ref{fig:workflow}, D$\rightarrow$F). Listing~\ref{lst:eca-rules} provides illustrative examples of accessibility-related risks that may trigger escalation, with thresholds that can be made profile-specific.

\begin{lstlisting}[caption={ECA rules for accessibility-driven triggers (conditions reference metric thresholds).},label={lst:eca-rules}]
RULE R1  // fluency over meaning
IF  Readability_FH > 80  AND  BERTScore < 0.85
THEN Activate HoTL supervision

RULE R2  // excessive deletion of essentials
IF  SARI_deletions > 0.40  AND  AlignScore < 0.80
THEN Activate HoTL supervision

RULE R3  // structural clarity compromised
IF  DSARI < theta_DSARI  OR  SAMSA < theta_SAMSA
THEN Activate HoTL supervision

\end{lstlisting}

Thresholds (e.g., 80 for readability, 0.85 for BERTScore) are calibrated per profile and domain, and can be revised as evidence is gathered. Additional rules may include violations of terminology/glossary constraints or missing mandatory sections, ensuring that escalation to HoTL remains sensitive to accessibility-critical failures.

\subsection{Key Performance Indicators (KPIs)}

KPIs are defined to connect evaluation with the user evidence in Section~4. Examples include comprehension gains from user tests, synonym-acceptance rates stratified by profile, glossary-activation rates in the UI, the recall--precision balance in CWI, and improvements in model-adaptation accuracy. 

To implement these indicators, ECA rules are specified in Listing~\ref{lst:kpis}. Listing 2 summarizes the KPI layer used for routing and monitoring. In addition, a Composite Quality Index (CQI) integrates Readability (0.4), Semantic Fidelity (0.3), and Structural Clarity (0.3), with weights summing to 1 and intended as an initial operational weighting for routing decisions. In the workflow (Figure~\ref{fig:workflow}), KPIs populate (E) and support monitoring, thresholding, and adaptation decisions together with rule checks.
\begin{lstlisting}[caption={Accessibility-oriented KPIs and composite quality index.},label={lst:kpis}]
KPI_1: IF comprehension_test_score(user) >= baseline + delta
       THEN record comprehension_gain

KPI_2: IF synonym_acceptance_rate(profile) >= theta_profile
       THEN validate lexical_clarity

KPI_3: IF glossary_activation_rate >= tau
       THEN confirm multimodal_support_used

KPI_4: IF recall_precision_balance(CWI) within [alpha, beta]
       THEN accept CWI_performance

KPI_5: IF model_adaptation_accuracy >= previous_cycle + epsilon
       THEN record adaptation_improvement

Composite_Quality_Index (CQI):
  CQI = 0.4*Readability + 0.3*Semantic_Fidelity + 0.3*Structural_Clarity

Decision threshold: CQI >= gamma (calibrated to comprehension outcomes)

\end{lstlisting}

\subsection{Integration of Evidence}

Beyond evaluation, the proposed KPIs and triggers can be implemented as structured preference signals for adaptation. While approaches such as Reinforcement Learning from Human Feedback (RLHF) \citep{Stiennon2020}, Direct Preference Optimization (DPO) \citep{Rafailov2023}, and other preference-based fine-tuning strategies \citep{Ziegler2019,Bohm2019}, including multi-agent feedback frameworks for adaptive simplification \citep{Mo2024}, do not strictly require domain-specific signals, accessibility-grounded KPIs provide a practical basis to ensure that adaptation is aligned with PL and ER principles rather than generic preferences. In this view, synonym-acceptance judgments, comprehension outcomes, and UI usability/usage signals extend beyond measurement and drive iterative improvement (Figure~\ref{fig:workflow}, F$\rightarrow$G$\rightarrow$B).

To make this actionable, what to measure (KPIs) and what to do (triggers) are encoded in a unified ECA policy that governs routing and adaptation; Listing~\ref{lst:eca-unified} abstracts both into a unified ECA policy.

\begin{lstlisting}[caption={Unified ECA formalization of KPIs and triggers.},label={lst:eca-unified}]
# KPI definition: measurable function of metrics and/or user signals
KPI_i = f(metrics, user_signals)

# Trigger: ECA rule referencing one or more KPIs
TRIGGER_j:
IF  KPI_i < theta_i  OR  combine(KPI_m, KPI_n) < threshold
THEN Activate HoTL supervision

\end{lstlisting}

\subsection{Illustrative Framework Execution}

To illustrate how the framework operates in practice, Table~\ref{tab:framework-example} presents a compact lexical simplification case derived from the user-study materials discussed in Section~4. The example shows how a potentially difficult term may motivate simplification, while still requiring escalation when a candidate substitutes risks changing the register or reducing domain adequacy.

\begin{table*}[t]
\centering
\small
\setlength{\tabcolsep}{5pt}
\renewcommand{\arraystretch}{1.15}
\begin{tabular}{p{0.16\textwidth} p{0.49\textwidth} p{0.25\textwidth}}
\hline
\textbf{Stage} & \textbf{Excerpt / signal} & \textbf{Framework interpretation} \\
\hline
Original excerpt &
``La caza y la pesca en la Comunidad de Madrid están sujetas a regulación especial y requieren una serie de \textbf{trámites}, entre ellos la obtención de licencia \ldots'' / \textit{``Hunting and fishing in the Community of Madrid are subject to special regulation and require a series of \textbf{administrative procedures}, including obtaining a licence \ldots''} &
The source text contains a potentially difficult domain-specific term in an administrative context. \\
\hline
Initial adapted output &
``\ldots requieren una serie de \textbf{papeleos} \ldots'' / \textit{``\ldots require a series of \textbf{paperwork} \ldots''} &
The output simplifies the term, but may introduce an overly colloquial register and reduce administrative precision. \\
\hline
Detected issue/trigger &
Readability may improve at the lexical level, but adequacy risk remains for domain register and terminology. / \textit{Possible mismatch between lexical simplicity and administrative appropriateness.} &
Trigger rule routes the case to HoTL review because lexical simplification alone does not guarantee appropriate wording. \\
\hline
HoTL-guided revision &
Review the candidate substitute against the checklist dimensions for terminology, clarity, and domain adequacy. / \textit{Check whether the simplified term remains precise, understandable, and appropriate for the target context.} &
Checklist dimensions affected: lexical clarity, terminology consistency, and appropriateness of register. \\
\hline
Approved output &
``\ldots requieren varios \textbf{trámites administrativos}, entre ellos obtener una licencia \ldots'' / \textit{``\ldots require several \textbf{administrative procedures}, including obtaining a license \ldots''} &
Meaning-preserving simplification approved; the wording is clearer while maintaining administrative precision. \\
\hline
\end{tabular}
\caption{Illustrative execution of the proposed framework on a lexical simplification case. Spanish excerpts are followed by an English gloss in italics.}
\label{tab:framework-example}
\end{table*}

This case illustrates a typical accessibility-sensitive situation for the proposed framework. The original term (\textit{trámites}) may motivate lexical simplification for some readers, but not every candidate substitute is equally appropriate. A form such as \textit{papeleos} may appear simpler, yet it introduces a more colloquial register and weakens the administrative precision and pragmatic appropriateness of the original wording in this document type. In such cases, automatic simplification may improve local readability while still requiring HoTL review to verify domain adequacy, terminology consistency, pragmatic fit, and appropriateness for the intended user profile. The revised version therefore preserves accessible wording without sacrificing the intended administrative meaning.

\subsection{Multi-agent Workflow Development}

As a practical implementation pathway, the framework could be developed as a multi-agent workflow, with specialized components for evidence retrieval, generation, evaluation, rule checking, KPI tracking, and human review. In this view, an orchestrator coordinates the interaction between generation, automatic assessment, escalation rules, and HoTL supervision, while preserving the traceability of intermediate decisions. Each agent performs a distinct role within the human-in/on-the-loop workflow, supporting traceability, explainability, and adaptive control. Taken together, these agents form a traceable and adaptive workflow in which automation and human supervision jointly support accessible, explainable, and accountable text generation. This architecture-specific realization is secondary to the present contribution, which remains the framework itself.

\section{Ethical Considerations and Governance}

The framework is designed for accountability. Every decision point is traceable via versioned prompts and policies, rule evaluations, KPI snapshots, and HiTL/HoTL logs, enabling post-hoc audit and model-update provenance. Explainability is supported by explicit ECA rules (why a route was chosen), checklist rationales (why edits were made), and profile-specific thresholds (why an item was approved or escalated). To protect users, logs follow data minimization and role-based access controls.

Governance safeguards ensure human oversight across the loop: HoTL sampling (5--10\% of outputs) applies even under auto-approve; it is mandatory after new model or policy releases; and auto-approve is disabled for high-risk domains (e.g., dosages in drug treatments appearing in a discharge summary report, legal warnings in contracts). These controls and profile-specific thresholds uphold oversight while enabling continuous, evidence-based adaptation.

This governance setup is consistent with the Ethics Guidelines for Trustworthy AI \citep{EC2019TrustworthyAI}, emphasizing human agency and oversight (HiTL/HoTL), transparency and explainability (machine-actionable rules with interpretable rationales), privacy and data governance (controlled retention and access), and fairness and accessibility (PL/ER principles). In addition, the proposed framework explicitly supports traceability and explainability as transversal safeguards: every generation, evaluation, and supervision step is logged and versioned, enabling end-to-end traceability from input prompts and human interventions to final outputs and model updates. Likewise, explainability is ensured by the declarative ECA rules and checklist rationales that make the decision path verifiable and intelligible both for experts and regulators. Together, these mechanisms reinforce accountability and foster confidence in the deployment of adaptive, human-supervised NLP systems for cognitive accessibility.

\section{Conclusions}

This paper proposes a hybrid framework that integrates Human-in-the-Loop (HiTL) and Human-on-the-Loop (HoTL) mechanisms to support cognitive accessibility in Plain Language and Easy-to-Read text generation. Grounded in empirical evidence from the EASIER project, expert annotation, and user interface validation, the framework translates accessibility principles into operational components such as checklists, trigger rules, and Key Performance Indicators (KPIs). These resources systematically complement automatic metrics with user-grounded evidence and expert supervision.

The framework builds on prior evidence suggesting complementary strengths across user groups, broader coverage from older adults and higher precision from people with intellectual disabilities, as well as the stabilizing role of expert annotations and the added value of interface validation for real-world deployment.
Moreover, integrating accessibility-based corpora and evaluation signals illustrates how human feedback can guide prompt design, fine-tuning, and preference-based alignment strategies.

By linking accessibility KPIs and triggers with adaptation methods such as RLHF and DPO, the framework connects evaluation and training into a traceable, explainable, and reproducible workflow. These mechanisms strengthen accountability and transparency, ensuring that accessibility standards are embedded not only in model outputs but also in the generation and supervision processes.

Future work will extend validation across domains, explore quantitative measures of comprehension gain, and develop the multi-agent orchestration framework, expanding the roles of evidence, evaluation, and reviewer agents to automate monitoring, explainability, and adaptive retraining. This direction aims to contribute to the standardization of accessibility-aware evaluation pipelines and to the creation of inclusive, transparent, and trustworthy NLP systems.

\section{Acknowledgements}
This work has been supported by grant \texttt{PID2023-1485770B-C21} (Human-Centered AI: User-Driven Adapted Language Models--HUMAN\_AI) by MICIU/AEI/\texttt{10.13039/501100011033} and by FEDER/UE.

\section{Bibliographical References}
\label{sec:reference}
\bibliographystyle{lrec2026-natbib}
\bibliography{paper}

\section{Language Resource References}
\label{lr:ref}
\bibliographystylelanguageresource{lrec2026-natbib}
\bibliographylanguageresource{languageresource}
A first set of reusable artefacts associated with the framework (checklist schema, ECA trigger templates, KPI definitions) is available through Zenodo \citep{HULATZenodo2026}.

\appendix
\section{An Illustrative Lexical Simplification Example}

This appendix provides the full passage underlying the illustrative example discussed in Table~\ref{tab:framework-example}. The example is included to clarify how automatic checks, escalation rules, and HoTL review interact in the proposed workflow. It is not presented as a new evaluation result, but as a compact demonstration of framework execution on a lexical simplification case derived from the materials used in the user studies described in Section~4.

\subsection{Original passage}

``La caza y la pesca en la Comunidad de Madrid están sujetas a regulación especial y requieren una serie de \textbf{trámites}, entre ellos la obtención de licencia, para poder practicarlas.'' / \textit{``Hunting and fishing in the Community of Madrid are subject to special regulation and require a series of \textbf{administrative procedures}, including obtaining a licence, in order to be practiced.''}

\subsection{Initial adapted version}

``La caza y la pesca en la Comunidad de Madrid tienen normas especiales y requieren una serie de \textbf{papeleos} para poder practicarlas.'' / \textit{``Hunting and fishing in the Community of Madrid have special rules and require a series of \textbf{paperwork} in order to be practiced.''}

\subsection{Illustrative revision after HoTL review}

``La caza y la pesca en la Comunidad de Madrid tienen normas especiales y requieren varios \textbf{trámites administrativos}, entre ellos obtener una licencia, para poder practicarlas.'' / \textit{``Hunting and fishing in the Community of Madrid have special rules and require several \textbf{administrative procedures}, including obtaining a licence, in order to be practiced.''}

\end{document}